\pdfoutput=1
\documentclass[abstract=true]{scrartcl}
\usepackage{lineno,hyperref}
\modulolinenumbers[5]

\usepackage[normal, bf]{caption}  
\usepackage[numbers,sort&compress]{natbib}

\usepackage{makeidx}  
\usepackage{amsmath}
\usepackage{amssymb}
\usepackage{subfig}
\usepackage{graphics} 
\usepackage{epsfig} 
\usepackage{tikz}
\usepackage{float}
\usetikzlibrary{arrows}
\usepackage{color}
\usepackage[english]{babel}
\usepackage[T1]{fontenc}
\usepackage{algorithmicx}
\usepackage{algorithm}
\usepackage{algpseudocode}

\usepackage{array}
\usepackage{booktabs}

\usepackage{enumitem}
\usepackage[framemethod=TikZ]{mdframed}
\newmdenv[leftmargin=0pt, innerleftmargin=0pt, rightmargin=0pt, innerrightmargin=0pt, outerlinewidth=0pt, linewidth=0pt,skipabove=10pt,skipbelow=0pt,frametitlerule=true,nobreak=true]{entry}

\newcommand{\lk}[1]{\hyperlink{#1}{#1}}

\newcommand{\info}[4]{#1 \textnormal{(\hypertarget{#2}{#2}) \\ \citeauthor{#3}, \citeyear{#3}, \cite{#3}, #4}}

\begin{document}
	
\title{Mitigating Metaphors: \\A \textit{Comprehensible} Guide to Recent Nature-Inspired Algorithms\footnote{Citation: Lones, M. A., Mitigating Metaphors: A Comprehensible Guide to Recent Nature-Inspired Algorithms, SN Computer Science 1, 49 (2020). This is the author's own version. The published version is available at DOI: \href{https://doi.org/10.1007/s42979-019-0050-8}{10.1007/s42979-019-0050-8}.}}

\author{Michael A. Lones\footnote{School of Mathematical and Computer Sciences, Heriot-Watt University, Edinburgh, UK}}

\date{}
\maketitle

\vspace{-5mm}
\begin{abstract}
In recent years, a plethora of new metaheuristic algorithms have explored different sources of inspiration within the biological and natural worlds. This nature-inspired approach to algorithm design has been widely criticised. A notable issue is the tendency for authors to use terminology that is derived from the domain of inspiration, rather than the broader domains of metaheuristics and optimisation. This makes it difficult to both comprehend how these algorithms work and understand their relationships to other metaheuristics. This paper attempts to address this issue, at least to some extent, by providing accessible descriptions of the most cited nature-inspired algorithms published in the last twenty years. It also discusses commonalities between these algorithms and more classical nature-inspired metaheuristics such as evolutionary algorithms and particle swarm optimisation, and finishes with a discussion of future directions for the field.
\end{abstract}

\section{Introduction}
This paper is intended to be an objective guide to the most popular nature-inspired optimisation algorithms published since the year 2000, measured by citation count. It is not the first paper to review this area \citep{yang2010nature, fister2013brief, xing16roughguide}, but it is arguably the first to present these algorithms in terms that will be familiar to the broader optimisation, metaheuristics, evolutionary computation, and swarm computing communities. Unlike some previous reviews, it does not aim to advocate for this area of research or provide support for the idea of designing algorithms based upon observations of natural systems. It only aims to report and summarise what already exists in more accessible terms.

The aim of this paper is not to explicitly criticise these approaches; other authors have already done this for nature-inspired metaheuristics in general \citep{sorensen2015metaheuristics} and for specific nature-inspired algorithms \citep{vcrepinvsek2012note, weyland15critical}. However, it is important to be aware of one point of criticism that was raised by \citep{sorensen2015metaheuristics}. This is the tendency for authors to present their algorithm from the perspective of, and using the terminology of, the domain of inspiration. Often nature-inspired algorithm papers begin with an initial review of a natural domain, then abstract this into a model of the domain, and this leads to an algorithmic description that contains terms from the domain. In many cases, this includes the introduction of new terms to describe well-established concepts from metaheuristics and optimisation. The consequence of this is that it can take considerable time and effort to understand how these algorithms work, even if the reader has a background in metaheuristics.

Well over a hundred nature-inspired algorithms have been published since 2000. For instance, the review book by \citet{xing16roughguide} names 134 of these, and the \textit{Evolutionary Computation Bestiary} \cite{bestiary} currently lists over 200. The premise for developing new  algorithms is often based solely on the desire to capture a behaviour observed in nature, with the assumption (rightly or wrongly) that it will also be relevant within an optimisation context. In more recent papers, it has become common to mention the No Free Lunch theorem \cite{wolpert1997no, joyce2018review} as a motivation. This theorem states that no optimiser is better than any other when considered across all possible optimisation problems, which can be interpreted as suggesting a need for diverse optimisers in order to solve diverse problems. Whether this is a valid assumption for the range of real world problems that optimisers are applied to in practice is unclear. Nevertheless, different optimisers are known to perform well on different problems, so there is some value to this argument.

These algorithms have gained a significant uptake. This can be seen in their citation counts: the 32 algorithms reviewed in this paper each have more than 200 citations; a third of them have more than 1000 citations. Given that most computer science papers achieve only a handful of citations per year, this is quite an achievement for a group of papers with an average age of around nine years. However, this combination of high uptake and opaque descriptions has led to fragmentation between the nature-inspired optimisation community and the wider metaheuristics community. To raise an observation that should be familiar to these communities: a certain amount of diversification is generally a good thing, but diversification without intensification tends to be ineffective. Applying this observation to the design of optimisation algorithms suggests that focusing on variants of a single nature-inspired algorithm (and most algorithms discussed in this paper do have a significant number of variants) is likely to be a sub-optimal approach and a potential waste of time and effort. This, in turn, suggests a need to tie back together these different threads of search. This paper aims to contribute towards this goal.

Section \ref{terms} presents the approach used to describe algorithms in this paper. Section \ref{algorithms} then uses this approach to describe the most widely-cited recent nature-inspired algorithms; the intent is for this to be used as a resource where someone can look up a particular algorithm and quickly gain an understanding of its main characteristics. Section \ref{novelty} then discusses the novelty of these algorithms in terms of both metaheuristic frameworks and broader metaheuristic concepts. Section \ref{pso} delves further into the specific overlaps between these algorithms and particle swarm optimisation and its variants. Section \ref{discussion} discusses some of the broader issues, and offers guidance on how research carried out in this area could be improved. Section \ref{conclusions} concludes.

\section{Descriptions and Terminology} \label{terms}

This paper attempts to describe algorithms using standard terms. However, this is not as straightforward as it may seem, since different parts of the metaheuristics community use different terminology. For example, those who work with local search algorithms refer to the transition between two points in the search space as a \textit{move}, and the result of evaluating a point is known as its \textit{objective value}. In the EA community, where much of the terminology derives from biological roots, these would be called \textit{mutation} and \textit{fitness}. In practice, both sets of terms are widely used. However, since the aim of this guide is to divorce nature-inspired algorithms from the terminology of their domain, generic terms will be used wherever possible, i.e. move rather than mutation.

When describing population-based algorithms, a further difficulty is that some algorithms are more naturally described using process-centric terms and others using population-centric terms. Particle swarm optimisation (PSO), for example, is essentially a distributed algorithm, and is easiest to present in terms of interactions between search processes. Genetic algorithms (GA), on the other hand, involve population-level operations such as selection; although these could be described as interactions between search processes, this would be convoluted and would make the algorithm harder to understand. Hence, in this paper a mixture of process-centric and population-centric terminology is used, depending on whether the algorithm is most appropriately described as the former or the latter.

In general, an attempt has been made to keep descriptions succinct and generic, whilst avoiding the definition of new terms. Little or no reference is made to an algorithm's source of inspiration from nature, unless this is required to understand the algorithm. Descriptions are intended to be sufficient to indicate the general characteristics of the algorithm, and to allow the reader to draw out similarities with other algorithms. They are not intended to be exhaustive, and hence some of the less important, or less specific, details are omitted. For example, consider the following description of PSO:

\vspace{4mm}
\begin{entry}[frametitle=\info{Particle Swarm Optimisation}{PSO}{eberhart1995particle}{$>$50000 citations}]
	Each search process has a velocity within the search space, and carries out moves by adding this to its current position at each iteration. The velocity is initially random. Then, at each iteration, each search process modifies its velocity by adding weighted terms based on the vector difference between its current point of search and the best points seen by both itself and by a subset of other search processes. This causes intensification of search by moving towards regions of the search space known to contain points of relatively high objective value. Diversification is provided by moving through the region between the current point and these target regions, and by overshooting these regions due to the momentum gained by maintaining a proportion of the existing velocity at each update.
\end{entry}
\vspace{4mm}

\noindent Unlike other presentations of this algorithm, this description does not use the terms \textit{particle} or \textit{informant}, since both of these can be described using generic terms. It does not go into detail about the exact form of each term in the velocity update equation, or how informants are allocated, since these details are not required to understand how the algorithm works, or how it relates to other algorithms. They are also subject to wide variation between implementations. The term \textit{velocity}, however, is used, since it is a well-defined concept within a vector space, and helps to understand the behaviour of the algorithm. The description also highlights algorithmic features which are expected to promote intensification and diversification of search.

A GA can be described as follows:
\vspace{3mm}
\begin{entry}[frametitle=\info{Genetic Algorithm}{GA}{holland1975ga}{$>$60000 citations}]
	At each iteration, search points with relatively high objective value are selected from the existing population. These are organised into pairs, and new search points are then created by exchanging solution components within pairs. This tends to sample the search space between existing search points, thereby both intensifying and diversifying search. Further diversification is provided by carrying out a random move away from the resulting points. Search points with relatively low objective value are removed from the population at each iteration, further intensifying search.
\end{entry}
\vspace{4mm}

\noindent This description is intentionally generic, since the exact details of how selection, recombination, mutation and solution replacement are implemented vary considerably. Domain-derived terms like crossover and mutation are not used, though \textit{select} is, since this has a clear non-domain meaning.

\section{Algorithms from A--Z} \label{algorithms}

\begin{figure}[tb!]
	\centering
	\includegraphics[width=  \columnwidth, trim={0 0 0.5cm 1.5cm},clip]{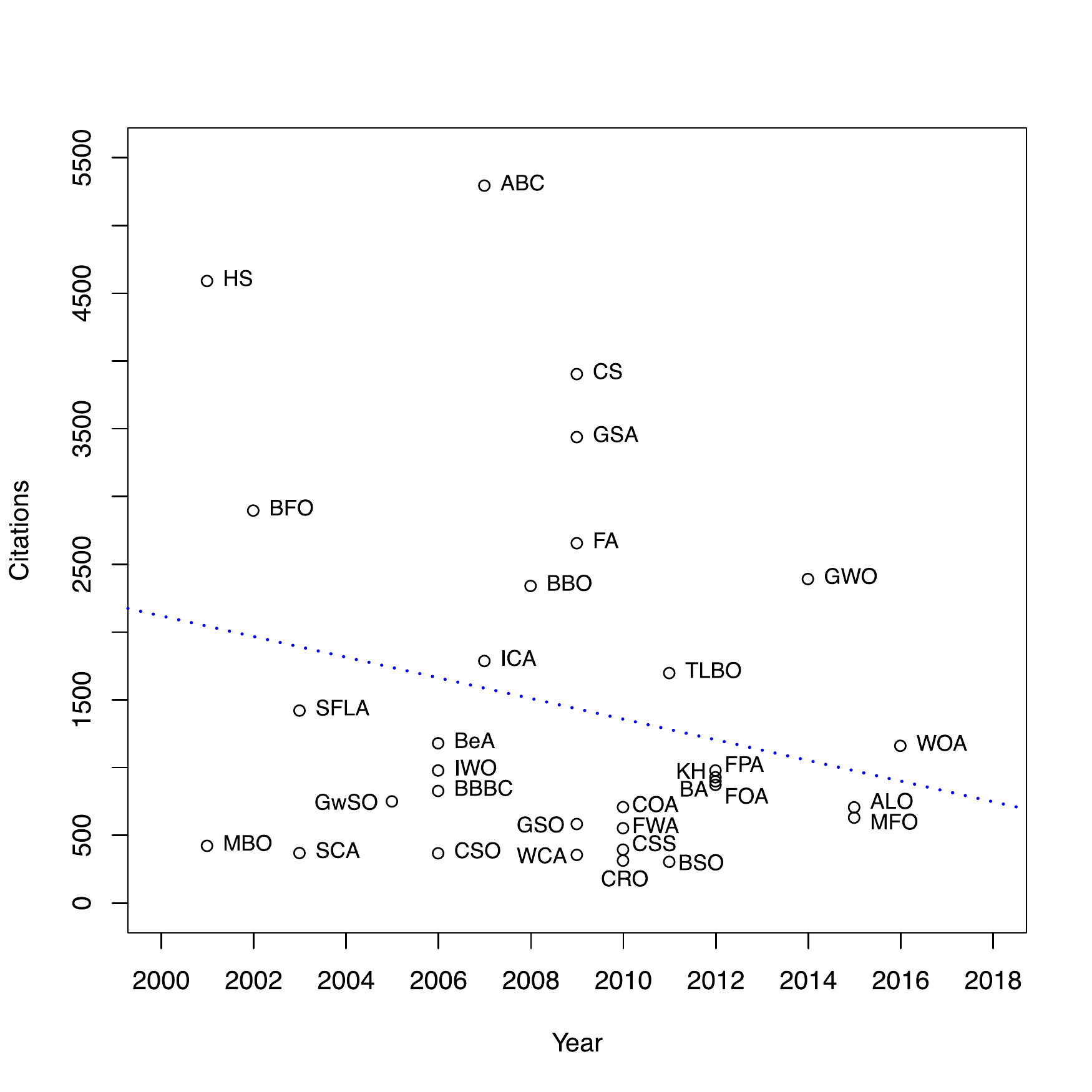}
	\caption{Citation counts of seminal papers, labelled with each algorithm's acronym. See Section \ref{algorithms} for full names and descriptions.}
	\label{fig:citations}
\end{figure}

Given the opaque nature of a lot of these papers, it would be a challenging task to read through and understand all the nature-inspired optimisation algorithms that have been published in recent years. Perhaps in reflection of this, previous reviews have generally described these algorithms using their original authors' words, or have focused on the sources of inspiration rather than trying to understand and present their underlying metaheuristic mechanisms. By comparison, the presentation in this paper aims to be comprehensi\textbf{\textit{ble}} rather than comprehensi\textbf{\textit{ve}}. Consequently, it focuses on the more popular of these algorithms. Popularity is measured in terms of citation count; this is not, of course, a robust measure of uptake, but it gives some indication of whether the algorithm has been used in practice. To bring the list of algorithms down to a manageable level, this review only covers those which have at least 200 citations, as measured by Google Scholar\footnote{Citations counts were collected in October 2019.}. By comparison, the seminal genetic algorithm (GA) work has $\sim$60000 citations, particle swarm optimization (PSO) has $\sim$50000 citations, ant colony optimization (ACO) has $\sim$10000 citations, and evolution strategies (ES) have $\sim$5000. It is notable that a number of algorithms in the list have citation counts approaching that of ES and collectively they have $\sim$30000 citations, roughly halfway between the citation counts of ACO and PSO. So, even taking into account the limitations of citation counts, they are clearly having an impact within the scientific record, and this alone should justify efforts to document and understand them. Fig. \ref{fig:citations} plots the approximate number of citations against the year that an algorithm's seminal paper was published. It can be seen from the trend line that the average citation count per year is $\sim$100.

The glossary below gives an overview of these 32 algorithms. Unless indicated otherwise, it is assumed that each algorithm is a population-based optimiser which updates the population synchronously over a period of iterations and begins with a population that is uniformly sampled from the search space.
\vspace{2.5mm}

\begin{entry}[frametitle=\info{Ant Lion Optimizer}{ALO}{mirjalili2015alo}{$>$300 citations}]
	At each iteration, search points with relatively high objective value are selected from the existing population. Search processes with relatively low fitness are then restarted within hyper-spherical regions centred around the selected points. Over time, the radius of the hyper-spheres is reduced, further intensifying search. The restarted search processes carry out random walks within their hyper-spherical regions; this is the main source of diversification within the algorithm. 
\end{entry}

\vspace{-2mm}
\begin{entry}[frametitle=\info{Artificial Bee Colony Algorithm}{ABC}{karaboga05abc}{$>$4500 citations}]
	Each search process generates local moves in the direction of the current position of another, randomly selected, search process. Only improving moves are accepted. The number of local moves generated by a particular search process is determined by the relative objective value of their current search point. Move sizes are probabilistic and are progressively reduced over time, leading to increased intensification. Diversification is promoted by restarting search processes which have not made progress within a certain number of moves at a randomly sampled location.
\end{entry}

\vspace{-2mm}
\begin{entry}[frametitle=\info{Bacterial Foraging Optimization}{BFO}{passino02bfo}{$>$2500 citations}]
	\begin{description}[topsep=0cm,leftmargin=0cm,itemsep=0pt,font=\normalfont\itshape]
		Each search process carries out sequential moves in the same direction until this no longer leads to improvement. When the current direction of search is no longer productive, a random change in direction occurs. After each iteration, the search processes with relatively low fitness are restarted at the current positions of search processes with relatively high fitness, intensifying search. The objective values of search points are adjusted by a crowding term, whose effect is to draw the search processes towards one another, further intensifying search.
	\end{description}
\end{entry}

\vspace{-2mm}
\begin{entry}[frametitle=\info{Bat Algorithm}{BA}{yang12ba}{$>$600 citations}]
	Search processes move towards the best solution within the population at different velocities, the magnitude of which is varied randomly at each iteration. There is also a probability of intensifying search by changing each search process's current position to a random point near the best solution within the population, with the likelihood of doing this decreasing each time a search process makes an improvement. The new search point is then accepted probabilistically, with a likelihood that increases each time a search process makes an improvement. 
\end{entry}

\vspace{-2mm}
\begin{entry}[frametitle=\info{Bees Algorithm}{BeA}{pham06ba}{$>$1000 citations}]
	At each iteration, solutions are randomly sampled within a fixed radius of the best solutions within the existing population. This radius reduces progressively over time and only improved solutions are accepted. Random restarts are used to maintain diversity. \textit{This is similar to an ES with a time-dependent mutation strategy.}
\end{entry}

\vspace{-2mm}
\begin{entry}[frametitle=\info{Big Bang-Big Crunch}{BB-BC}{erol06bbbc}{$>$600 citations}]
	At the start of each iteration, a point representing the objective value-weighted average of the previous population is calculated and a new population is created by sampling from a normal distribution centred around this point. The width of the distribution is reduced at each iteration, intensifying search. \textit{This can be seen as an estimation of distribution algorithm (EDA) with a simple generative model.}
\end{entry}

\vspace{-2mm}
\begin{entry}[frametitle=\info{Biogeography-Based Optimizer}{BBO}{simon08bbo}{$\sim$2000 citations}]
	At each iteration, for each solution in the population, components are replaced by copying them from other solutions; the likelihood of this is proportional to the solution's objective value, and the likelihood of choosing another solution as a source is proportional to its objective value. To promote diversity, a local move is carried out away from each resulting solution, with the probability of doing so inversely proportional to its objective value. \textit{This has similarities to a multi-parent GA.}
\end{entry}

\vspace{-2mm}
\begin{entry}[frametitle=\info{Brain Storm Optimization}{BSO}{shi2011brain}{$>$300 citations}]
	At each iteration, the population is clustered using k-means clustering, and the best solution in each cluster is identified. Each solution in the population is then considered for replacement by comparing its objective value against that of a new search point and then keeping the best. Most of the time, this new search point is generated by either a local move from an existing solution, or by recombining two existing solutions in a GA-like manner. In either case, selection of the existing solution(s) is biased towards the cluster bests. To diversify the population, there is also a mechanism to sample random solutions during this process.
\end{entry}

\vspace{-2mm}
\begin{entry}[frametitle=\info{Cat Swarm Optimization}{CSO}{chu06cat}{$\sim$300 citations}]
	At each iteration, each search process either carries out a local search, or moves in the direction of the best solution in the population. When carrying out a local search, a specified number of points are sampled in the vicinity of the current position and the best one is kept.
\end{entry}

\vspace{-2mm}
\begin{entry}[frametitle=\info{Charged System Search}{CSS}{kaveh10css}{$\sim$600 citations}]
	All search processes carry out moves towards the current positions of other search processes. Search processes have velocities, and the speed with which a search process moves towards a given search process is calculated using an inverse-square law weighted by the objective value of that process's current point of search. Adaptive parameter changes allow the degree of attraction to vary over time, and the best solutions within the population are always preserved. \textit{Note that this algorithm is similar to FA and GSA.}
\end{entry}

\vspace{-2mm}
\begin{entry}[frametitle=\info{Chemical Reaction Optimization}{CRO}{lam10cro}{$>$300 citations}]
	Search processes carry out either local search or a more disruptive global search using disruptive operators (such as the GA recombination operator). The balance between local and global search, and the likelihood of accepting non-improving solutions, are both based on the history of the population member: if no improvement has been made for a while, global search replaces local search; if solutions with lower objective values were previously accepted, then they are less likely to be accepted in the future. \textit{This algorithm has similarities to memetic algorithms and simulated annealing.}
\end{entry}

\vspace{-2mm}
\begin{entry}[frametitle=\info{Cuckoo Optimization Algorithm}{COA}{rajabioun11coa}{$\sim$500 citations}]
	At each iteration, new search points are sampled within a radius of each existing search point, and only the best points are kept. The resulting population is then clustered using k-means clustering, and the cluster with the highest mean objective value is identified. Search points in the other clusters are then moved towards the fittest cluster.
\end{entry}

\vspace{-2mm}
\begin{entry}[frametitle=\info{Cuckoo Search}{CS}{yang09cs}{$\sim$3000 citations}]
	Uses a small population of solutions. At each iteration, a search process with a relatively low objective value is restarted, either at a randomly sampled location, or by applying a `L\'{e}vy flight' to another, randomly selected, solution. L\'{e}vy flights are a kind of random walk with step sizes generated from a heavy-tailed probability distribution.
\end{entry}

\vspace{-2mm}
\begin{entry}[frametitle=\info{Firefly Algorithm}{FA}{yang09fa}{$>$2000 citations}]
	At each iteration, all search processes carry out moves towards the current positions of other search processes. The degree of movement towards each point is calculated using an inverse-square law weighted by its objective value, causing intensification of search towards points with higher objective values. \textit{Note that this algorithm is similar to CSS and GSA.}
\end{entry}

\begin{entry}[frametitle=\info{Firework Algorithm}{FWA}{tan10fwa}{$>$300 citations}]
	At each iteration, solutions are sampled in a neighbourhood around the best solutions in the population. Only improving moves are accepted. Neighbourhoods are sampled using a Gaussian distribution centred around the current point. The width of the distribution is inversely proportional to the objective value of the best solution, causing increased intensification as search progresses.
\end{entry}

\begin{entry}[frametitle=\info{Flower Pollination Algorithm}{FPA}{yang12fpa}{$>$500 citations}]
	At each iteration, search processes carry out moves either towards the best solution in the population or the current position of a randomly selected search process. In the former case, the step size is determined by sampling a L\'{e}vy distribution (see CS).
\end{entry}

\begin{entry}[frametitle=\info{Fruit Fly Optimization Algorithm}{FOA}{pan12foa}{$>$600 citations}]
	All search processes carry out moves towards the best solution in the population. However, how this is achieved is unclear from the description.
\end{entry}

\begin{entry}[frametitle=\info{Glowworm Swarm Optimization}{GwSO}{krishnanand05gso}{$>$600 citations}]
	Each search process maintains a numerical value that summarises its recent search progress, increasing this when it finds a an improving search point and decreasing gradually when it makes no progress. At each iteration, each search process carries out moves towards another search process located within a hyper-spherical region centred around its current point; a search process with a high search progress value is more likely to be chosen as a target, and the radius of this region shrinks when there are many search processes nearby. Citation count includes \citep{krishnanand09gso}.
\end{entry}

\begin{entry}[frametitle=\info{Gravitational Search Algorithm}{GSA}{rashedi09gsa}{$>$2500 citations}]
	All search processes carry out moves towards the current positions of other search processes. Search processes have velocities, and the speed with which a search process moves towards a given search process is calculated using an inverse-square law weighted by the objective value of that process's current point of search. \textit{Note that this algorithm is similar to CSS and GSA.}
\end{entry}

\begin{entry}[frametitle=\info{Grey Wolf Optimizer}{GWO}{mirjalili14gwo}{$>$1000 citations}]
	At each iteration, each search process carried out moves around the edges of a hypercube centred around a target search point. The target point is selected from a region bounded by the three current best search points within the population. Hypercubes become gradually smaller at each iteration in order to intensify search, and there is a random component in the update equation to inject diversity.
\end{entry}

\begin{entry}[frametitle=\info{Group Search Optimizer}{GSO}{he09gso}{$>$500 citations}]
	The search processes with the current best solutions carry out local moves, using a mathematical model of animal vision to delimit the region they explore at a particular time. The majority of the other search processes carry out moves towards the search processes with the current best solutions. The remaining search processes generate diversity within the population by carrying out random walks.
\end{entry}

\begin{entry}[frametitle=\info{Harmony Search}{HS}{geem01hs}{$\sim$4000 citations}]
	At each iteration, a single new solution is created from a randomly selected existing solution. For each of its decision variables, a new value is chosen either at random or by copying and slightly modifying the value from another randomly-selected solution. If the new solution has a higher objective value than the worst solution in the population, it replaces it. \textit{Note that this algorithm has been proven equivalent to a form of ES \citep{weyland15critical}.}
\end{entry}
\begin{entry}[frametitle=\info{Imperialist Competitive Algorithm}{ICA}{atashpaz07ica}{$\sim$1500 citations}]
	A population is randomly initialised and the best solutions are selected. For each of these solutions, a sub-population is created with size proportional to its objective value and is filled randomly using the remaining search points within the population. At each iteration, the solutions in the sub-population are moved towards the best solution within the sub-population, with some noise added to inject diversity. Then, each sub-population is given a value based mainly upon the objective value of its best solution, and solutions in sub-populations with low values are re-allocated to sub-populations with high values. The algorithm terminates when there is a single non-empty sub-population.
\end{entry}


\vspace{-2.5mm}
\begin{entry}[frametitle=\info{Invasive Weed Optimization}{IWO}{mehrabian06iwo}{$>$750 citations}]
	At each iteration, search processes sample a number of local moves from their current position. The number of moves is proportional to the relative objective value of their current position, and the size of moves decreases non-linearly over time. Once the population size reaches an upper bound, search processes with relatively poor solutions are ended, and the solutions they generated are removed from the population.
\end{entry}

\begin{entry}[frametitle=\info{Krill Herd}{KH}{gandomi12kh}{$>$600 citations}]
	Search processes carry out moves towards the population best, their historical best, and the objective value-weighted average of the population. They also carry out moves towards or away from search processes within a given radius, based upon their objective value. Search processes also have a component of random movement. The weighting of components is time-dependent, with less random motion and more movement towards the population best as time proceeds. On top of this, GA-like operations are carried out within the population.
\end{entry}

\begin{entry}[frametitle=\info{Marriage in Honey Bees Optimization}{MBO}{abbass01mbo}{$\sim$400 citations}]
	At each iteration, a number of random walks are carried out, starting from the locations of the best solutions in the population. New solutions are created using an operator that recombines the existing (start of walk) solution with solutions encountered during the walk. The likelihood of this occurring at each step of the walk is based on objective value, and also reduces over the course of the walk. Move sizes progressively decrease during the walk. Local search is used to improve solutions at each iteration of the algorithm; the operator used for this is chosen probabilistically based on its past success rate.
\end{entry}

\begin{entry}[frametitle=\info{Moth-Flame Optimization}{MFO}{mirjalili15mfo}{$\sim$250 citations}]
	Search processes carry out moves in a spiral path towards a target point. The target points are the historical best solutions of other search processes. Initially, all historical bests are used as targets, with the particle that has the highest current objective value moving towards the highest historical best, and the search process with the least current objective value moving towards the lowest historical best. Over time, fewer targets are followed.
\end{entry}

\vspace{-2mm}
\begin{entry}[frametitle=\info{Shuffled Frog Leaping Algorithm}{SFLA}{eusuff03sfla}{$>$1000 citations}]
	At each iteration, the population is split into sub-populations, each with a broad objective value spread. Each sub-population is then repeatedly sub-sampled by objective value, and the worst solution in each sub-sample is moved towards the best solution in the sub-sample (or alternatively the population). In each case, if this does not lead to improvement, the solution is replaced by a random search point. After each sub-population has been processed, the sub-populations are merged, and the procedure is repeated.
\end{entry}

\vspace{-2mm}
\begin{entry}[frametitle=\info{Society and Civilisation Algorithm}{SCA}{ray03sca}{$>$300 citations}]
	At each iteration, the population is clustered. In each cluster, the best solutions are selected. The remaining solutions in the cluster are then moved towards the selected solutions. A similar procedure is then carried out for the selected solutions from all clusters, with the worst solutions amongst these moved towards the best solutions. The algorithm also takes into account constraint satisfaction.
\end{entry}

\vspace{-2mm}
\begin{entry}[frametitle=\info{Teacher-Learning Based Optimization}{TLBO}{rao11tlbo}{$>$1000 citations}]
	At each iteration, the mean position of the population is calculated and subtracted from the population's best search point. Moves are then carried out by adding a fraction of the resulting vector to each population member \textit{(this is similar to differential evolution)}. Only improving moves are accepted. Each population member is then compared to another randomly selected member; if the target has a higher objective value, it is moved towards it; otherwise, it is moved away. Again, only improving moves are accepted.
\end{entry}

\vspace{-2mm}
\begin{entry}[frametitle=\info{Water Cycle Algorithm}{WCA}{shah09iwd}{$\sim$250 citations}]
	At each iteration, population members with relatively high objective value (but not the population best) are moved closer to the population best by a random amount. The remaining population members are each moved closer to one of these relatively high objective value solutions by a random amount, with proportionally more of them moving towards the best solutions. Random local moves are also applied to maintain diversity.
\end{entry}

\vspace{-2mm}
\begin{entry}[frametitle=\info{Whale Optimization Algorithm}{WOA}{mirjalili16woa}{$\sim$250 citations}]
	Each search process carries out moves in a hypercube around a target search point and iteratively moves towards this target either by shrinking the hypercube or through a spiral motion. Target choice is affected by a time-dependent parameter; initially this causes random members of the population to be followed; later all search points follow the population best.
\end{entry}

\section{Commonalities} \label{novelty}

Recent nature-inspired metaheuristics have sometimes been criticised for a lack of novelty. Before discussing this in more detail, it is first useful to consider the meaning of the term \textit{metaheuristic}. Many authors who develop nature-inspired algorithms use this term as a synonym for ``optimisation algorithm'', but this is not the original meaning of the term, which is more akin to a generative model that can be used to guide the development of a particular algorithm.  \citet{sorensen2018history} address this disparity by distinguishing metaheuristic algorithms (i.e.\ particular implementations of a metaheuristic idea) from metaheuristic frameworks (i.e.\ the more general models from which these algorithms are derived). This distinction is important when talking about novelty, because whilst there is considerable scope for designing a novel metaheuristic algorithm, there is much less scope for developing a novel metaheuristic framework. For instance, you can create a novel metaheuristic algorithm by modifying the mutation operator used by a GA, or by hybridising a GA with an operator from PSO, but in both cases there is no novel metaheuristic framework being created. It is worth noting that hybridisation, in particular, introduces combinatorial scope for generating algorithms that are technically novel, yet which introduce no novel algorithmic features.

Whilst metaheuristic frameworks are a useful concept for narrowing the definition of novelty, it can also be useful to talk about recurring ideas that appear within multiple frameworks. For instance, EAs and PSO are probably good candidates for being called metaheuristic frameworks, but there are clearly common concepts that occur within both of these; for example, the way in which both techniques have mechanisms for exploring search points that are intermediate to existing ones. In a previous paper \citep{lones2014metaheuristics}, an attempt was made to identify and describe some of these more general metaheuristic approaches; an abridged listing of these is reproduced in Table \ref{table:metaheuristics}.

\begin{table*}[h!]
	\centering
	\begin{tabular}{>{\raggedright}p{3.1cm}p{6cm}>{\raggedright\arraybackslash}p{4cm}}
		\toprule
		\textbf{Concept} & \textbf{Description} & \textbf{Examples} \\
		\midrule
		Hill Climbing
		& Follow a sequence of local improvements to reach a locally optimal solution.
		& Steepest ascent, stochastic hill climbing \\ \midrule
		Accepting Negative Moves
		& Allow moves to worse solutions.
		& Threshold accepting, simulated annealing \\ \midrule
		Restarts 
		& Restart the search process in a different region once it has converged at a local optimum.
		& Random-restart hill climbing, iterated local search\\ \midrule
		Adaptive Memory Programming
		& Use memory of past search experience to guide future search.
		& Tabu search, EAs, PSO\\ \midrule
		Population-Based Search
		& Multiple cooperating search processes that run in parallel.
		& EAs, PSO, scatter search\\ \midrule
		Intermediate Search
		& Explore the region between two or more previously visited search points.
		& Crossover, PSO, path relinking\\ \midrule
		Directional Search
		& Identify productive directions within the search space, and carry out moves accordingly.
		& Gradient ascent, CMA-ES, PSO\\ \midrule
		Variable Neighbourhood Search
		& Search different neighbourhoods around the location of a known local optimum.
		& PSO, variable neighbourhood search\\ \midrule
		Search Space Mapping
		& Construct a map to guide search processes that are traversing the search space.
		& ACO, guided local search, DIRECT\\ 
		\bottomrule
	\end{tabular}
	\caption{A list of recurring metaheuristic concepts. Adapted from \citep{lones2014metaheuristics}.}
	\label{table:metaheuristics}
\end{table*}

Technically, almost all the algorithms described in the previous section meet the definition of a novel metaheuristic algorithm, since they differ from standard metaheuristic algorithms such as ESs, GAs and standard PSO. However, it is difficult to argue that any of them are novel metaheuristic frameworks, since most of them clearly borrow (or perhaps re-discover) concepts that are also central to conventional metaheuristic frameworks. Referring to the metaheuristic concepts listed in Table \ref{table:metaheuristics}, all of the algorithms described in the previous section implement a combination of \textit{hill climbing}, \textit{adaptive memory programming} and \textit{population-based search}, and this is also true of EAs and PSO. The majority also implement some form of \textit{intermediate search}, most commonly using either a PSO-like operator that picks a point geometrically between two existing points or an EA-like crossover operator that recombines solution components. Those which use PSO-like operators also carry out \textit{directional search} in a similar manner to PSO. Many of the algorithms use \textit{restarts} (\lk{ABC}, \lk{BFO}, \lk{BeA}, \lk{CS}, \lk{SFLA}), which are also commonly used in local search algorithms. Many also have strategies for \textit{accepting negative moves}: some of these resemble simulated annealing (\lk{BA}, \lk{CRO}); however, the most common approach involves random walks (\lk{ALO}, \lk{BFO}, \lk{CS}, \lk{GSO}, \lk{KH}, \lk{MBO}), which might be considered a degenerate form of threshold acceptance, but is otherwise a relatively novel idea. Several algorithms use search trajectories that follow a spiral-like path around local optima (\lk{GWO}, \lk{MFO}, \lk{WOA}), and this could be considered a form of \textit{variable neighbourhood search}. 

In terms of resemblance to existing metaheuristic frameworks, a large proportion of the algorithms have a clear resemblance to PSO in that a population of search processes move towards each other using vector-based operations (\lk{ABC}, \lk{BeA}, \lk{BA}, \lk{COA}, \lk{CSO}, \lk{CSS}, \lk{FA}, \lk{FOA}, \lk{FPA}, \lk{GSA}, \lk{GSO}, \lk{GWO}, \lk{GwSO}, \lk{KH}, \lk{MFO}, \lk{TLBO}, \lk{WCA}, \lk{WOA}). 
A few algorithms might be considered variants of ES (\lk{BeA}, \lk{HS}, \lk{IWO}), and a number of algorithms are broadly EA-like (\lk{BBO}, \lk{BSO}, \lk{COA}, \lk{ICA}, \lk{SFLA}, \lk{SCA}), with a number of these hybridising PSO-like operators (\lk{COA}, \lk{ICA}, \lk{SFLA}, \lk{SCA}). Some algorithms have notable degrees of self-similarity: for instance, \lk{CSS}, \lk{FA} and \lk{GSA} all use inverse-square laws to calculate the attraction between search processes.

\section{Commonalities with PSO} \label{pso}

PSO has clearly been an influence to many of the nature-inspired algorithms reviewed in Section \ref{algorithms}, and consequently it is important to dig down further to understand how ideas explored within this group of algorithms intersect with those explored in the PSO literature.

The majority of PSO-style algorithms listed in Section \ref{algorithms} have similar basic mechanics to PSO, in that search processes move towards other search processes using vector operations. A major difference is that the majority of these algorithms (all except \lk{KH} and \lk{MFO}) do not use historical bests, i.e. the best point of search seen by a particular search process. This means that search processes are influenced only by the current locations of both themselves and other search processes. The metaheuristic motivation for this is unclear, since it appears to reduce the amount of information available to guide search. Nevertheless, it should be noted that the idea of ``social only'' interactions (i.e. ignoring a search process's own search experience) has been explored in PSO and in both \citep{kennedy1997particle} and \citep{pedersen2010simplifying} was found to have no significant effect upon the algorithm's performance when applied to certain problems; however, this is not the same as not recording historical bests, since search processes are still influenced by the historical bests of other search processes.

Another major difference from standard PSO is that most of the algorithms have no direct analogue of velocity or momentum; rather, move sizes are determined using simpler rules, including time-dependent move sizes (\lk{ABC}, \lk{ALO}, \lk{BB-BC}, \lk{CSS}, \lk{IWO}, \lk{MBO}), distance dependent move sizes (\lk{CSS}, \lk{FA}, \lk{GSA}) and region-based sampling (\lk{ALO}, \lk{BA}, \lk{BeA}, \lk{BB-BC}, \lk{FWA}, \lk{GWO}, \lk{WOA}). Time dependent move sizes have also been explored in variants of PSO \citep{shi1999empirical, ratnaweera2004self}. Region-based sampling involves directly sampling from a region of search space that is shaped or bounded by one or more search points, rather than applying vector-based operations. This approach has earlier been used in Bare Bones PSO \citep{kennedy2003bare}, where it was introduced as a means of simplifying the dynamics of PSO and making it more tractable for analysis. Distance-dependent move sizes are notable: usually in PSO, search processes move faster towards informants that are further away, meaning that move size increases with distance. In \lk{CSS}, \lk{FA} and \lk{GSA}, on the other hand, the search processes are less influenced by distant search processes, so move size reduces with distance. This causes interactions between search processes to become geographically localised, which could be useful for multi-modal landscapes; however, it is unclear whether the resulting behaviour is more effective than other mechanisms introduced to PSO to handle these kind of landscapes, such as multi-swarm approaches \citep{blackwell2004multi}.

A consequence of using simpler update rules is that the dynamics of many of these algorithms are much simpler than in standard PSO. A benefit of this is that it potentially makes their behaviour easier to understand. However, by removing exploratory dynamics like overshooting and oscillation, there is a danger that they will only explore the regions between existing search points and suffer premature convergence as a result. To address this, most include one or more mechanisms to promote diversification. These include hybridisation with local search (\lk{CSO}, \lk{CRO}, \lk{COA}, \lk{FWA}, \lk{IWO}, \lk{MBO}, \lk{WCA}), random restarts (\lk{ABC}, \lk{BFO}, \lk{BeA}, \lk{CS}, \lk{SFLA}), random walks (\lk{ALO}, \lk{BFO}, \lk{CS}, \lk{GSO}, \lk{KH}, \lk{MBO}) and spiral-like movements (\lk{GWO}, \lk{MFO}, \lk{WOA}). The latter, in particular, may lead to search trajectories that resemble those seen in PSO (and it should be noted that a similar approach is used in spiral optimisation \citep{tamura11spiral}). Hybridisation with local search is also fairly common in PSO, e.g. \citep{chen05particle}, where restarts have also been used \citep{kaucic13multi}. Random walks are arguably one of the more interesting mechanisms explored in recent nature-inspired optimisation algorithms, particularly those that build upon biological knowledge in this area, e.g. \lk{CS} and \lk{BFO}, and there is no real analogue in the PSO literature.

When carrying out moves towards other search processes, the manner of choosing target search points varies widely amongst the algorithms in the list. Some (\lk{BA}, \lk{CSO}, \lk{FOA}) only use the population best, relying on other mechanisms (e.g. restarts) to maintain diversity. Several algorithms (\lk{GwSO}, \lk{WCA}, \lk{GWO}, \lk{COA}) choose targets in a fitness-informed manner, either probabilistically, by selecting the top \textit{n} solutions in the population, or in the case of \lk{COA}, by clustering and picking the cluster with the highest mean fitness. These approaches are somewhat related to variants of PSO that use dynamic allocation of informants, e.g. \citep{du15adequate}. A number of algorithms have mechanisms that cause particles to be more influenced by nearby search processes. This includes those that relate move size to distance (see above). It also includes \lk{SCA}, which dynamically clusters the population based on distance. Distance-based selection of informants has also been used in PSO, e.g. \citep{lane08particle}. A number of algorithms use all other search processes as targets, either directly (\lk{CSS}, \lk{FA}, \lk{GAO}), or indirectly by summarising information about them (\lk{KH}, \lk{BB-BC}). Similar ideas have been investigated in variants of PSO, such as fully-informed PSO \citep{mendes04fully} and quantum PSO \cite{sun2004global}. Some algorithms use time-varying rules for choosing targets, notably those that move from randomly-chosen targets towards the population best over time (\lk{MFO}, \lk{WOA}).
The idea of dynamically-varying the number of informants over time has also been explored in the PSO literature \citep{suganthan99particle}.

\section{Discussion} \label{discussion}

Are recent nature-inspired algorithms \textit{novel}? Yes and no. On the one hand, most (but certainly not all) of the algorithms reviewed in this paper are distinct from existing optimisation algorithms, and given a particular search space, they would likely follow different trajectories to existing algorithms. On the other hand, many of these algorithms use variants of well-established metaheuristic concepts that are also found in existing metaheuristic frameworks such as PSO, EAs and local search. Furthermore, the analysis of PSO-style algorithms shows that many of their underlying ideas have also been explored by the more mainstream PSO community. However, chronologically, this hasn't always been in one direction. Sometimes the PSO community has explored these ideas earlier, sometimes later, and sometimes in parallel to recent nature-inspired algorithms. Either way, it shows how the fragmentation of the nature-inspired computing community has led to duplicated effort.

Are recent nature-inspired algorithms \textit{competitive}? This is less clear. Most of the cited papers include a performance evaluation. The results are not reported here, because almost all show the algorithm to perform better than the algorithms they were compared against. Even without taking No Free Lunch theorems \citep{wolpert1997no} into account, it is implausible to believe that this is true for all of them. This is not to say that the results are incorrect, but it does reflect the difficulty of designing fair comparative studies \citep{piotrowski15regarding, fong16recent, vcrepinvsek2016comparison, garcia2017since}. We can speculate that all of these algorithms will \textit{sometimes} perform better on \textit{some} problems when compared against other algorithms, since problem landscapes are diverse, and small differences in the topography of a landscape can favour different approaches.

However, given a \textit{specific} problem, it is difficult to know which algorithm will work well. The field of meta-learning \citep{lemke2015metalearning} has been studying this issue for some time, but progress on understanding how problems can be characterised, categorised and mapped to specific optimisers has so far been limited. This means that performance on one problem currently tells us little about potential performance on another problem, and consequently that practitioners usually have to try out a range of different optimisers to determine which one works well on their problem. In a sense, the recent developments in nature-inspired algorithms have increased the number of optimisers available to try out. This may sometimes be beneficial, but it also makes it harder for a practitioner to identify a suitable optimiser that is well-understood and has community support. Given the vast scope for creating variants and hybrids of existing algorithms, this situation is only likely to get worse.

An alternative, and arguably more promising, direction of travel can be seen in the hyperheuristics \citep{burke2013hyper, epitropakis2018hyper} and broader machine learning communities \citep{li17learningtooptimize, wichrowska17learnedoptimizers}. Both address the problem of choosing an optimiser as an optimisation problem, using a machine learning algorithm to identify an optimiser that is good at solving a specific task. In the case of hyperheuristics, the optimiser, which is usually an evolutionary algorithm, can be used to construct new optimisation algorithms. This can be done either by specialising an existing algorithm (for example, evolving a new mutation operator for an EA) or by assembling existing metaheuristic components in a novel way. In effect, the latter is an automated version of the many manual attempts to hybridise metaheuristics that can be found in the literature. However, this automated approach is currently limited by a lack of standardised interfaces \citep{swan15mhinthelarge}, and this arguably is limited by the tendency of the community to think of metaheuristics in terms of algorithms rather than re-usable components.

This focus on algorithms rather than components is a particular issue for the nature-inspired algorithm community, where the objective of domain modelling is almost always the generation of a single algorithm that captures all pertinent behaviours present within the domain of inspiration. As a consequence, any interesting, novel, components extracted from the domain tend to become conflated with other, less interesting, and sometimes arbitrary, components. This makes it hard to understand the relevance and contribution of individual components within the optimisation setting. Arguably a better approach would be to identify any component of the domain that is particularly interesting, and integrate this individually within one or more existing metaheuristic frameworks. Even better would be to make the code available in re-usable form: it could then be used by other algorithm developers, or even used as a new building-block within hyperheuristic frameworks.

An important barrier that stands in the way of this kind of integration is the success of previous authors who have not followed this path. This can be seen in the large citation counts amongst recent nature-inspired optimisation papers, and the initial career boost that this may provide to their authors. It is perhaps less apparent that association with a part of the field that is seen as less rigorous may result in a career penalty in the long run. Many of the citations to these papers come from researchers who work in applied optimisation. This brings up another important factor in the success of nature-inspired optimisers, the false assumption that new means better, which leads to inexperienced practitioners using the most recent, rather than the most appropriate, metaheuristic to solve a particular optimisation problem. This is a difficult problem to address, because it is arguably caused by sociological rather than technological factors, and (due to the wide reach of optimisation) spans a broad range of academic communities. Nevertheless, maybe efforts, such as this, to tie together the loose ends of the community may contribute towards a solution.

\section{Conclusions} \label{conclusions}

Numerous papers describing new nature-inspired optimisation algorithms have been written over the past 20 years. Unfortunately, it has become common practice for these papers to describe algorithms using non-standard terminology derived from their domain of inspiration, resulting in papers that are often very difficult to read and understand. In this paper, an attempt has been made to describe the most widely cited of these algorithms using standard metaheuristic terminology. It is hoped that the resulting descriptions will make it easier for readers to gain a quick understanding of how these algorithms work, without having to read the original papers. As a result, this should make it more straightforward for metaheuristics practitioners to read, review and understand work that uses these algorithms.

This paper also makes an attempt to analyse the commonalities between algorithms. The existing literature has raised particular cases where there is a strong similarity between different nature-inspired optimisation algorithms, but opaque terminology makes it hard to recognise these similarities in general. The standardised descriptions in this paper make this process easier, and this has been demonstrated by relating each of the algorithms to existing metaheuristic concepts. The resulting analysis suggests that few of the algorithms introduced in the last 20 years introduce fundamentally new concepts; rather, they mostly reassemble existing concepts in new ways. Since many of the algorithms are swarm-like, a closer look was taken at their commonalities with particle swarm optimisation and its variants. This revealed few points of absolute novelty, suggesting that the two communities have largely been following the same tracks. Perhaps suprisingly, it was noted that particle swarm optimisation did not always get there first.

This paper also emphasises the need to bring together the different threads of the metaheuristic community, with the aim of reducing redundancy, making research results more accessible, and developing new approaches that integrate the diverse work that is being done. Some important work has already started in this area, including, for example, efforts to standardise interfaces between metaheuristic components. However, arguably a lot more effort is required if we are to reduce the fragmentation of the field and leverage the diverse talents of the metaheuristics community in useful ways.


\bibliographystyle{abbrvnat}
\bibliography{mitigatingmetaphors}

\end{document}